\documentclass{article}

\usepackage{times}
\usepackage{graphicx} 
\usepackage{subfigure} 

\usepackage{natbib}
\usepackage{multirow}

\usepackage{algorithm}
\usepackage{algorithmic}

\usepackage{hyperref}

\usepackage[accepted]{icml2016}

\icmltitlerunning{Can Evolutionary Sampling Improve Bagged Ensembles?}

\begin{document} 

\twocolumn[
\icmltitle{Can Evolutionary Sampling Improve Bagged Ensembles?}

\icmlauthor{Harsh Nisar}{nisar.harsh@gmail.com}
\icmlauthor{Bhanu Pratap Singh}{bhanupratap.mnit@gmail.com}
\icmlkeywords{machine learning, evolutionary algorithm}

\vskip 0.3in
]

\begin{abstract} 
Perturb and Combine (P\&C) group of methods generate multiple versions of the predictor by perturbing the training set or construction and then combining them into a single predictor \cite{breiman1996bias}. The motive is to improve the accuracy in unstable classification and regression methods. One of the most well known method in this group is Bagging. Arcing or Adaptive Resampling and Combining methods like AdaBoost are smarter variants of P\&C methods. In this extended abstract, we lay the groundwork for a new family of methods under the P\&C umbrella, known as Evolutionary Sampling (ES). We employ Evolutionary algorithms to suggest smarter sampling in both the feature space (sub-spaces) as well as training samples. We discuss multiple fitness functions to assess ensembles and empirically compare our performance against randomized sampling of training data and feature subspaces.
\end{abstract} 

\section{Introduction}
Bagging and various variants of it have been widely popular and studied extensively in the last two decades \cite{breiman2001random,breiman1996bagging,breiman1999pasting,ho1998random}. There has been notable work in understanding the theoretical underpinning of bootstrap aggregating and as to what makes it such a powerful method \cite{domingos1997does,buchlmann2002analyzing}. In traditional bagging, each training example is sampled with replacement and with probability $\frac{1}{N}$. Adaptive Resampling and Combining (Arcing) techniques which modify the probability of each training example being sampled based on heuristics have also been developed and widely used \cite{freund1996experiments,breiman1999using}. 

Random subspace methods also known as attribute bagging refer to creating ensembles of predictors trained on randomly selected subsets of total features, that is, predictors constructed on randomly chosen sub-spaces.

Both methods, sub-sampling and sub-spacing reduce the variance of the final ensemble and hence increase the accuracy. Arcing methods are known to reduce the bias of the model as well.   

Error based resampling algorithms which try to set the train-set error to zero \cite{freund1996experiments}, designed bagged ensembles with minimal intersection \cite{papakonstantinou2014bagging}, diversity and uncorrelated errors \cite{kuncheva2003measures,tang2006analysis}, importance sampling \cite{breiman1999pasting} etc. are some of the areas being studied to improve bagged ensembles. Either there are multiple answers to the question, or the answer changes with each dataset.

Instead of figuring out precisely as to \textit{what} sampling and combination of training sets make a bagged ensemble better, we try to fix the \textit{definition} of better, and allow the bootstrapped training sets to \textit{evolve} themselves in order to align with the definition. We generate multiple sampled candidate training sets for the final ensemble and let them compete, mutate and mate their way to the optimal sampling and combination. Evolutionary computation has been used for selection of different predictors to be part of an ensemble \cite{gagne2007ensemble} and also for the selection of the most suitable machine learning pipeline for a classification problem \cite{Olson2016EvoBIO}.

To our best knowledge, genetic algorithms haven't been directly used to evolve bootstrapped samples of the training data. 
\vspace{-1.2em}
\begin{table}[h]
\centering
\caption{Genetic algorithm's experiment setting}
\label{table:ga_parameter}
\vspace{0.1em}
\scalebox{0.8}{
\begin{tabular}{ll}
\hline
GA parameter                  & Value                              \\ \hline
Selection                     & 3-way tournament\\
Crossover                     & 2-point crossover                  \\
Population Size               & 30                                 \\
Per-individual mutation rate  & 0.4                                \\
Per-individual crossover rate & 0.6                                \\
Generations                   & 30                                \\
Ensemble size in an individual     & 10                      \\           \hline
\end{tabular}}
\end{table}
\vspace{-2.2em}
\begin{table*}[t]
\centering
\caption{\small{Results comparing performance of first individual (FI) and hall of fame (HOF) on unseen data. Values are averages (standard deviations) over 100 runs. Statistical tests are p-values of paired t-tests on the test mean squared error (mse) compared between FI and HOI. Win(\%) is the proportion of runs where HOF had a lower mse than FI on unseen data.}}
\label{table:results}
\vspace{0.2em}
\scalebox{0.7}{
\begin{tabular}{ll|l|lll|ll}
\hline
         &                       & &\multicolumn{3}{c|}{\textit{Sub-sampling}} & \multicolumn{2}{c}{\textit{Sub-spacing}}      \\ \hline
Data Set & & Parameters             & FEMPO        & FEMPT         & FEGT         & FEMPO         & FEGT         \\ \hline \hline
\textbf{Servo}&    & HOF - Mean (Std. Dev.) & 0.16 (0.11)  & 0.2 (0.43)    & 0.26 (0.23)  & 0.25(0.29)    & 0.59(0.52)   \\
Nr. Inputs & 4         & FI - Mean (Std. Dev.)  & 0.20 (0.15)  & 0.2 (0.38)    & 0.28 (0.32)  & 0.71(0.63)    & 0.72(0.49)   \\
\#Training & 167 & Win (\%)               & 64           & 43            & 52           & 94            & 65           \\
\#Test & 10\% & p-value                & 0            & 0.475         & 0.47         & 0             & 0.01         \\ \hline
\textbf{Ozone} &   & HOF - Mean (Std. Dev.) & 18.20 (4.77) & 18.28 (6.02)  & 18.79 (5.62) & 18.48 (5.38)  & 19.11(5.98)  \\
Nr. Inputs & 8          & FI - Mean (Std. Dev.)  & 18.12 (4.58) & 18.15 (5.62)  & 19.66 (6.55) & 20.16 (5.59)  & 20.27(6.02)  \\
\#Training & 330  & Win (\%)               & 51           & 52            & 57           & 67            & 62           \\
\#Test & 10\% & p-value                & 0.71         & 0.59          & 0.025        & 0             & 0.01         \\ \hline
\textbf{Boston} &  & HOF - Mean (Std. Dev.) & 12.49 (6.88) & 13.09 (7.09)  & 12.74 (5.4)  & 20.78 (11.35) & 21.02(12.23) \\
Nr. Inputs & 12          & FI - Mean (Std. Dev.)  & 12.96 (6.76) & 13.11 (6.6)   & 13.95 (6.96) & 25.07 (13.41) & 27.8(15.06)  \\
\#Training & 506 & Win (\%)               & 63           & 48            & 58           & 75            & 81           \\
\#Test & 10\% & p-value                & 0.16         & 0.93          & 0            & 0             & 0            \\ \hline
\textbf{Abalone} &  & HOF - Mean (Std. Dev.) & 4.93 (0.37)  & 5.025 (0.402) &    4.81(0.26)          & 5.09(0.59)    & 5.02(0.54)   \\
Nr. Inputs & 8  & FI - Mean (Std. Dev.)  & 4.92 (0.35)  & 5.012 (0.409) &       4.86(0.22)       & 5.49(0.71)    & 5.4(0.62)    \\
\#Training & 4177 & Win (\%)               & 54           & 40            &      72        & 90            & 84           \\
\#Test & 25\% & p-value                & 0.42         & 0.25          &     0         & 0             & 0           \\ \hline
\end{tabular}
}
\end{table*}

\section{Algorithm}
Evolutionary computation techniques evolve a population of solution variables (bootstrapped training sets in our case) to optimize towards a  given criteria. The fittest offspring across all the generations is considered as the most optimal solution. In Evolutionary Sampling (ES), we followed a standard Genetic Algorithm. Initially, a population of multiple ensembles is generated by randomly sampling from the training data multiple times.  Each ensemble (henceforth referred to as an individual) in a generation is evaluated based on a fitness function. Fit individuals are selected for the next generation. After this, crossover is applied on a fixed percentage of individuals wherein two individuals swap their predictors. Post this, a fixed percentage of individuals unaffected by crossover undergo a random mutation. Randomly selected member datasets from the selected individual have some of their rows/features deleted, replaced or inserted with equal probability. In feature sub-spacing the features are subject to perturbation whereas in sub-sampling rows are perturbed.   

We've used the Python package DEAP \cite{DEAP_JMLR2012} to implement ES. GA parameters are shown in table \ref{table:ga_parameter}. As suggested before, instead of understanding as to what makes a bagged ensemble better, we try to rely on the definition of better and try to evolve our ensemble into the same. The fitness function is what guides the sampling and combination of the different sampled datasets. We propose three fitness functions and then try to analyse their performance.

\textit{FEMPO}: Fitness Each Model Private Out of Bag. It takes each predictor part of the candidate ensemble and measures their performance on the samples that were left out of it's training bag \cite{breiman1996out}. Final fitness of the ensemble is the mean of each model's RMSE.

\textit{FEMPT}: Fitness Each Model Private Test.  It is the average of the performance of member predictors on a private test-set which is held out for each sampled dataset during its instantiation. 

\textit{FEGT}: Fitness Ensemble Global Test. During the start of the algorithm, 20\% of the training data is set aside. Each ensemble's prediction is based on the average prediction of it's member predictors. RMSE is calculated against the set aside global test.

\section{Experiment}
We conduct experiments on two variants of sampling : Sub-sampling and Sub-spacing. Sub-sampling works on sampling training examples, whereas sub-spacing works on generating multiple feature sets. We conduct our experiments on 4 benchmark datasets [see table \ref{table:results}].  We compare the mean squared error of the first individual (FI) of the first generation with the hall of fame (best individual) after 30 generations. We assume that the first individual in the first generation is representative of an ensemble which randomly samples its rows or features like in traditional bagging. We try to analyse whether ES is able to evolve better ensembles starting from random specimens. We uniformly use an unpruned Decision Tree Regressor with max depth arbitrarily set as 5. 

\section{Results and conclusion}
A 50\% win-ratio would suggest that the performance of the ensemble after undergoing ES is better than its random counterpart only half the times. Mean MSE and standard deviation of the same is also a good metric to compare ES with random instantiation. 
The null hypothesis in the paired t-test suggests that the average mean squared error between the two methods is the same. If the p-value is smaller than a threshold, then we reject the null hypothesis of equal averages.

In sub-sampling, FEMPO and FEGT performs equally or better than their random counterparts. Though the win percentages are almost half in many cases, it could be that the algorithm was initialized with an optimal combination and sampling. FEGT shows the most improvement in Abalone (72\% win-ratio).

In sub-spacing, both FEMPO and FEGT do significantly better than random sub-spacing. GA has definitely helped in improving accuracy of the model. One should note, GA has a narrow exploration space in case of feature sub-spacing as compared to sub-sampling and features play a more significant role in deciding the model's behaviour than a few rows of data.

Results suggest that ES is possibly useful in cases where maximum accuracy needs to be juiced out and computation is not an issue.  It's evident that better and more robust fitness functions need to be explored, even multi-objective fitness functions,  which better represent generalizability and error of the ensemble.

It needs to be explored how these methods can be used to generate different models for smaller segments or patches of the dataset \cite{breiman1999pasting}. Can the segments, suggested by ES along with fitness functions that take into account each model’s fitness (FEPT or FEMPO), be used to find different cohorts in the dataset?

ES guided sub-spacing using linear base estimators can be useful in high dimensional problems like genomic data where selecting features is very important while keeping final models interpretable.

It will be interesting to see what happens if the algorithm is allowed to run for generations till the fitness test error reduces approximately to zero. We plan to experiment with different base estimators for each sampled dataset and also explore how sub-spacing and sub-sampling can be combined into one algorithm.

Going with the theme of reproduction, we've  released the basic framework for ES on GitHub (http://github.com/evoml/evoml). We encourage researchers to contribute to the project and test out different fitness functions themselves. 

\bibliographystyle{icml2016}

\end{document}